\pdfoutput=1
\documentclass[11pt]{article}
\usepackage[table]{xcolor}
\usepackage{META/emnlp2021}
\usepackage{times}
\usepackage{url}
\usepackage{latexsym}
\usepackage[T1]{fontenc}
\usepackage{calc}
\usepackage{tikz}
\usepackage{xkeyval}
\usepackage{ifthen}
\usepackage{hyperref}
\usepackage[normalem]{ulem}
\usepackage{soul}
\usepackage{multirow}
\usepackage{todonotes}
\useunder{\uline}{\ul}{}
\usepackage[utf8]{inputenc}
\usepackage{microtype}
\usepackage{xcolor}
\usepackage{geometry}
\usepackage{times}
\usepackage{latexsym} 

\title{A Comprehensive Comparison of Word Embeddings\\ in Event \& Entity Coreference Resolution.}
\author{Judicael POUMAY \\
  ULiege/HEC Liege\\
  Rue Louvrex 14, 4000 Liege, Belgium \\
  {\tt judicael.poumay@uliege.be} \\\And
  Ashwin ITTOO \\
  ULiege/HEC Liege\\
  Rue louvrex 14, 4000 Liege, Belgium \\
  {\tt ashwin.ittoo@uliege.be} \\}

\date{November 2020}
\def\checkmark{\tikz\fill[scale=0.4](0,.35) -- (.25,0) -- (.8,.7) -- (.25,.2) -- cycle;} 

\begin{document}
\maketitle

\begin{abstract}
Coreference Resolution is an important NLP task and most state-of-the-art methods rely on word embeddings for word representation. However, one issue that has been largely overlooked in literature is that of comparing  the performance of different embeddings across and within families in this task. 
Therefore, we frame our study in the context of Event and Entity Coreference Resolution (EvCR \& EnCR), and address two questions   : 
1) Is there a trade-off between performance (predictive   \&  run-time) and embedding size? 
2) How do the embeddings' performance compare within and across families?
Our experiments reveal several interesting findings. 
First, we observe diminishing returns in performance with respect to embedding size. 
E.g. a model using solely a character embedding achieves 86\% of the performance of the largest model (Elmo, GloVe, Character) while being 1.2\% of its size. 
Second, the larger model using multiple embeddings learns faster overall despite being slower per epoch. 
However, it is still slower at test time.
Finally, Elmo performs best on both EvCR and EnCR, while GloVe and FastText perform best in EvCR and EnCR respectively.
\end{abstract}
\section{Introduction} 	
Coreference Resolution (CR) is  an important NLP task.
It can be subdivided into Event and Entity Coreference Resolution (EvCR and EnCR).
These tasks serves as the basis for several downstream applications such as information extraction, text summarization, machine translation and text mining \cite{humphreys1997event,azzam1999using,werlen2017using,corefbio}.

State-of-the-art methods for CR\cite{barhom-etal-2019-revisiting,lee-etal-2017-end,joshi-etal-2019-bert} rely on various word embeddings for word representation.  
These embeddings are organized into three families: \textit{static}, \textit{contextual} and \textit{character} embeddings  \cite{staticreview,contexsurvey,charPOS}, each differing in size.
Contextual embeddings are larger (1024) compared to the other families (usually 300 for static and 50 for character). 
They also tend to outperform the other families in most tasks but lead to larger and heavier models \cite{devlin-etal-2019-bert,peters-etal-2018-deep}. 
We are thus confronted with a trade-off of performance (predictive  \&  run-time) vs. dimensionality. 
Moreover, embeddings also differ within families which also leads to differences in predictive performance.

Several studies investigated how different embeddings influence  the predictive performance in different tasks  \cite{berardi2015word,gromann2018comparing,joshi-etal-2019-bert,li2018comparison}. 
However, the two aforementioned issues of the performance vs. dimensionality trade-off and performance variations within and across embedding families have been overlooked to a large extent, especially in coreference resolution. 
Literature is still unclear about which embeddings perform best in which tasks, and whether larger, more expressive embeddings should also be preferred or whether some predictive performance can be compromised for improved run time. 

Thus, we seek to address two questions in the context of CR: 
1) Is there a trade-off between performance (predictive \& run-time) and embedding size? 
2) How do the embeddings' performance compare within and across families? 
The current state-of-the-art in EvCR \cite{barhom-etal-2019-revisiting} rely on three families of embeddings for word representation, and thus provides a suitable frameworks for addressing our research questions. 
Starting from the original model of \newcite{barhom-etal-2019-revisiting}, we performed various experiments and ablative studies across and within each family of embeddings, resulting in  16 different models. 
\footnote{The relatively large number of models and experiments is one reason why we preferred to focus on a single task}. 
We compared their predictive performance, size (number of parameters) , run-time and memory usage.

We discovered high level of diminishing returns in term of predictive performance per embedding. 
The smallest model (using solely a character embedding \cite{charPOS}) achieves 86\% of the performance of the largest model (GloVe \cite{pennington-etal-2014-glove} , ELMo \cite{peters-etal-2018-deep}, Character embedding) with 1.2\% of its size. 
Hence, incorporating additional embeddings leads to diminishing returns in terms of predictive performance.
In addition, we found that size and run-time are weakly correlated: larger (more complex) models can converge faster (number of epochs and total training time) than smaller ones . 
In terms of predictive performance, we found GloVe and FastText perform best in EvCR and EnCR respectively in their family with ELMo being the best overall. 
Moreover, we found that the smallest aforementioned model outperforms Word2Vec ($\sim$+10 F1), yielding predictive performance close to the previous state-of-the-art \cite{kenyon-dean-etal-2018-resolving} in EvCR (68.43 vs 69 F1). 
Our results can have important implications for practitioners in implementing CR and other NLP models in real-life applications. 
\section{Background and Related work} 	
\label{relwork}

\subsection{Word embeddings families}
Literature generally distinguishes between three families: \textit{static}, \textit{contextual} and \textit{character} embeddings  \cite{staticreview,contexsurvey,charPOS}.

\textit{Static embeddings}, such as word2vec, FastText, and GloVe, create a one-to-one mapping between words and their vector representations. 
Word2vec \cite{word2vec} learns through a language modelling task by either learning to predict a word given its context (CBOW) or predict the context given a word (Skip-gram). FastText \cite{bojanowski-etal-2017-enriching} learns sub-words embeddings which are then combined for each word. Finally, GloVe \cite{pennington-etal-2014-glove} relies on word co-occurrence information. Both Glove and FastText are trained on a  Skip-gram task.

\textit{Contextual embeddings} take into account the context of a given word, i.e. their vector representations changes depending on surrounding words.
ELMo is a Bi-LSTM trained on a language modelling task. 
GPT-2 is similar except that it is unidirectional.
Finally, BERT is based on a transformer architecture and trained on a masked language modelling task.

Lastly, \textit{character embeddings} learn vectors based on character sequences \cite{charPOS}. 

Since their development, word embeddings have been very largely studied \cite{tan-etal-2015-lexical,CHEN2018178,WANG201812,clark-etal-2019-bert,tenney-etal-2019-bert} and a complete literature review is out of the scope of our work. Hence, we will focus on studies closest to ours. First, we will review studies on embeddings' performance regardless of the task. Then, we move to our task of interest which is coreference resolution. 

\subsection{Studies on Embeddings' Performance}
\newcite{gromann2018comparing}  found that FastText (0.812 F1) outperformed Polyglot (0.675 F1) and Word2Vec (0.750 F1) for ontology alignment. They used two ontologies: Global Industry Classification Standard and Industry Classification Benchmark.
They also demonstrated the ability of FastText to better handle out-of-vocabulary words. 

\newcite{berardi2015word}  found that Word2Vec (Accuracy (ACC) 43.63\%) outperformed  polyglot  (ACC 4\%) and GloVe  (ACC 30.21\%) on a word analogy test using Wikipedia and a collection of Italian books (mostly novels) as datasets.  

\newcite{joshi-etal-2019-bert}  found that BERT  significantly outperformed ELMo on EnCR (+11.5 F1) on the GAP and OntoNotes datasets. 

\newcite{li2018comparison} found that GloVe outperformed FastText and Word2Vec on a tweet classification task, especially when trained on specific corpora, viz.CrisisLexT6, CrisisLexT26, and 2CTweets.

\subsection{Word embeddings in Coreference Resolution.}
\begin{figure}[th]
    \centering
    \includegraphics[width=\linewidth]{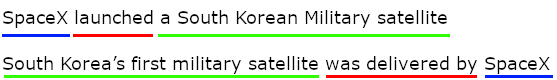}
    \caption{Two coreferent event mentions with colors indicating associated coreferent entity mentions.}
    \label{fig:CR_ex}
\end{figure}

Event Coreference Resolution and Entity Coreference Resolution (EvCR and EnCR respectively) are concerned with clustering Event and Entity mentions that refer to the same reality \cite{barhom-etal-2019-revisiting,lee-etal-2017-end}. Figure \ref{fig:CR_ex} depicts two event mentions with the same meaning. 

Events mentions refer to textual representations of real-life events. As can be seen from Figure \ref{fig:CR_ex}, events generally consist of a  trigger word (most often a verb), such as "launched", and a set of arguments, such as "SpaceX" and "a South Korean Military satellite".  Four argument types are generally distinguished: Arg0, Arg1, location, and time, as defined in \newcite{barhom-etal-2019-revisiting}, where Arg0 (resp. Arg1) is the closest entity on the left (resp. right) of the trigger word.
These arguments are optional and often referred to as entities. The goal of EvCR (and EnCR) is to identify which events (and  entities)  are coreferent with each other and to cluster them. 

We now briefly review studies using word embeddings for EnCR and EvCR.

\textbf{EnCR :}  \newcite{lee-etal-2017-end} used GloVe as word representation allied with a Bi-LSTM and attention mechanisms. Their model achieved state-of-the-art (68.8 F1) on the the CoNLL-2012 corpus. As already mentioned, \newcite{joshi-etal-2019-bert} reported higher EnCR performance when using BERT compared to ELMo: +3.9 F1 in OntoNotes and +11.5 F1 in GAP.  

\textbf{EvCR :} \newcite{choubey-huang-2017-event} relied on GloVe for EvCR using the ECB+ corpus \cite{ECBplus}. They used a joint modelling approach to perform within and cross document EvCR and achieved state-of-the-art performance. The same corpus was employed by \newcite{barhom-etal-2019-revisiting}, who proposed an EvCR/EnCR model based on  ELMo \cite{peters-etal-2018-deep}, GloVe \cite{pennington-etal-2014-glove} as well as a fine-tuned character embedding. Similarly, it jointly performs EnCR and EvCR.  Their model yielded performance of 79.5 F1 in EvCR.

\section{Methodology} 	
\label{modelsec}
\subsection{Original model}
\begin{figure}[h]
    \centering
    \includegraphics[width=\linewidth]{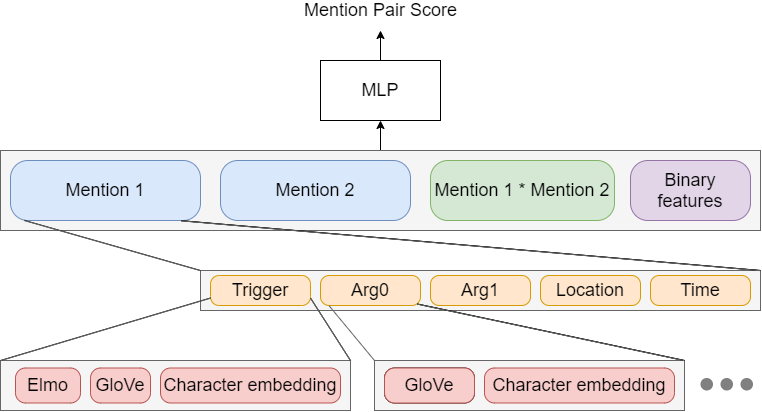}
    \caption{Original input structure of \newcite{barhom-etal-2019-revisiting}'s model.}
    \label{fig:input}
\end{figure}

Our approach is based on the state-of-the-art model of \newcite{barhom-etal-2019-revisiting}, which we refer to as the \textsc{Original} \footnote{\textsc{modelname} denotes a model} model. This model consists of two neural networks, which jointly resolve entities and events coreferences. Figure \ref{fig:input} shows the input of both networks. The two event (resp. entity) mentions embeddings are in blue and the green box represents an element-wise multiplication of the mentions. Finally, binary features indicate whether the two encoded mentions have coreferent arguments. The constituents of each mention, i.e. trigger, Arg0, Arg1, Location and time, are  represented by a static (GloVe) and a character embedding. The trigger is also represented by a contextual embedding (ELMo). Furthermore, the character embedding is fine tuned during training while the contextual and static embeddings are not.

The input dimensionality is 3*(1024+5*(300+50)) + 200 = 8522, where 1024, 300 and 50 are the dimensions of ELMo, GloVe and the character embeddings, and 200 corresponds to the size of the binary features.  This input is then fed into two subsequent ReLU layers with dimensions equal to half the input dimension (4261 neurons each). Since the number of parameters is proportional to the square of the input dimension, we have a model size exceeding 54 million parameters, computed as ($\frac{input^2}{2}+(\frac{input}{2})^2+\frac{input}{2}$).

\subsection{Derived models}
The gist of our methodology involves substituting and/or removing specific embeddings from \newcite{barhom-etal-2019-revisiting}'s original model (which uses 3 embeddings : static=GloVe, contextual=ELMo and character), resulting in 16 different models shown in Table \ref{tab:models}.  In the first group of models, one, two, or three (of  the three) embeddings are removed from the original model. In the second group, the static embedding is changed to Word2Vec (Skip-gram) or FastText (other embeddings are either left unchanged or removed). Similarly, in the third group the contextual embedding is changed to BERT or GPT-2 (other embeddings are either left unchanged or removed). Note: in Table \ref{tab:models}, gray rows denote identical models. 

We implemented our models using Pytorch. Models were trained and tested following \newcite{barhom-etal-2019-revisiting}'s procedure. Pre-trained vectors and models were used for the embeddings. Our code is available online
\footnote{\href{https://github.com/JudicaelPoumay/event_entity_coref_ecb_plus}{github.com/JudicaelPoumay/event\_entity\_coref\_ecb\_plus}}.

\begin{table}[ht]
\begin{tabular}{|l|l|l|l|}
\hline
\multicolumn{1}{|c|}{{\ul Model}} & \multicolumn{1}{c|}{{\ul Stat.}} & \multicolumn{1}{c|}{{\ul Ctx.}} & \multicolumn{1}{c|}{{\ul Char.}} \\ \hline
\multicolumn{4}{|c|}{\textit{\textbf{Group 1: Across family study}}} \\ \hline
  \rowcolor{lightgray}
Original \shortcite{barhom-etal-2019-revisiting} & GloVe  & ELMo & \checkmark \\ \hline
Contextual/Static & GloVe   & ELMo  & X      \\ \hline
Contextual/Char & X      & ELMo   & \checkmark     \\ \hline
Static/Char    & GloVe   & X    & \checkmark     \\ \hline
Static    & GloVe   & X    & X     \\ \hline
Contextual & X     & ELMo  & X     \\ \hline
Char            & X    & X     & \checkmark     \\ \hline
No word embed       & X    & X     & X      \\ \hline
\multicolumn{4}{|c|}{\textit{\textbf{Group 2: Within family study: Static}}} \\ \hline
  \rowcolor{lightgray}
GloVe           & GloVe & ELMo   & \checkmark     \\ \hline
Word2Vec        & Word2Vec & ELMo   & \checkmark     \\ \hline
FastText        & FastText  & ELMo  & \checkmark     \\ \hline
Only GloVe      & GloVe  & X    & X       \\ \hline
Only FastText   & Word2Vec & X     & X       \\ \hline
Only Word2Vec    & FastText & X    & X       \\ \hline
\multicolumn{4}{|c|}{\textit{\textbf{Group 3: Within family study: Contextual}}} \\ \hline
  \rowcolor{lightgray}
ELMo & GloVe & ELMo & \checkmark  \\ \hline
BERT & GloVe & BERT & \checkmark  \\ \hline
GPT-2& GloVe & GPT-2  & \checkmark  \\ \hline
Only ELMo   &   X& ELMo    & X       \\ \hline
Only BERT       &  X& BERT    & X       \\ \hline
Only GPT-2     &  X&  GPT-2  & X       \\ \hline
\end{tabular}
\caption{List of trained and tested model and their components. Ctx. = Contextual; Stat. = Static; Char. = Character; X/\checkmark indicate absence/presence of an input.} 
\label{tab:models}
\end{table}

\section{Experimentation setup}
\label{evalset}

\subsection{Dataset}
The dataset we use for our study is ECB+ \cite{ECBplus}. Together with EECB \cite{lee-etal-2012-joint}, it is one of the largest datasets for within and cross document EvCR and EnCR \cite{lee-etal-2012-joint,barhom-etal-2019-revisiting}.
Both EECB and ECB+ are extensions of ECB \cite{bejan-harabagiu-2010-unsupervised} and consist of English Google News documents clustered into topics and annotated for coreference. For more details on the ECB+ corpus statistics, please refer to \newcite{barhom-etal-2019-revisiting}.

Other dataset for coreference resolution exist : GAP, OntoNotes, CoNLL 2012, ACE,  TAC KBP and MUC. However, the definition of coreference resolution in these corpora do not suits our study and model. For example, GAP is a corpus  of  ambiguous  pronoun-name pairs while ECB+ defines mentions cluster for events and their entities \cite{joshi-etal-2019-bert}. OntoNotes annotates coreferences but does not indicate which mentions is an event and which is an entity. MUC, ACE, and TAC KBP do not provide cross document coreferences\cite{ijcai2018-773}. Finally, while CoNLL 2012 defines an event coreference task, events represent only a small portion of the all the coreferent mentions and again it does not provide cross document coreferences \cite{pradhan2012conll}. In-depth reviews of the listed datasets are provided in  \cite{STYLIANOU2021114466, ijcai2018-773, sukthanker2018anaphora}.

\subsection{Experiments}

We performed three sets of experiments.
The first set concerns models of Group 1  (see Table \ref{tab:models}). We investigated the impact of removing one, two, or three (of the three) embeddings from the original model. Our aim was  to determine the contribution of the different embeddings (static, contextual and character) on the predictive performance of the \textsc{Original} model. Thus, the models will have varying sizes, translating into varying run-time and memory requirements. Therefore, for this set of experiments, we also report on model size (number of parameters), run-time (seconds) and memory usage (RAM). 

The second (third) set concerns models of Group 2 (Group 3) (see Table \ref{tab:models}) and aim at investigating the contributions of static (contextual) embeddings.

For the latter two experiments, we do not consider model size as all possible sizes would have been investigated in group 1.
For all experiments, we will report the predictive performance achieved by the various models with the CoNLL F1 and MUC F1  metrics \cite{coreferenceMeasures}.

Following \newcite{barhom-etal-2019-revisiting}'s original paper, we can claim that a difference of 1 point between any two models is significant with a p-value $<$ 0.001. This confirms that our results are statistically sound and not due to randomness.

\section{Results}
\label{results}
\subsection{Results 1: All Embedding Families }
\label{ablasec}
\begin{figure*}[ht]
    \centering
    \includegraphics[width=\linewidth]{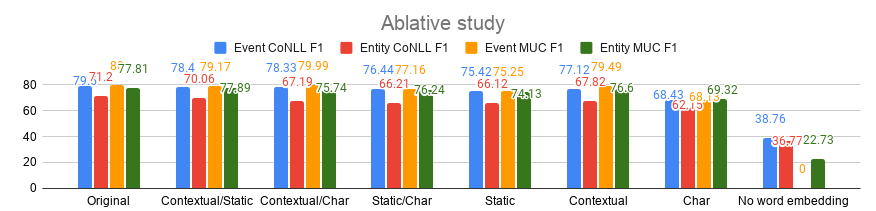}
    \caption{Comparing the predictive performance of the original model (using 3 embeddings) with models where we removed one, two or all three embeddings.}
    \label{fig:ablanal}
\end{figure*}

As mentioned earlier, our aim was to investigate the contributions of the static (Glove) , contextual (ELMo) and character embedding to the original model's performance via an ablative study. The predictive performance scores (CoNLL/MUC F1) of Group 1 models are in Figure \ref{fig:ablanal}, respectively from left to right. 

A first observation is that the baseline performance differs between the two measures (CoNLL \& MUC F1). This is due to the mention identification effect \cite{coreferenceMeasures} which makes CoNLL F1 more optimistic than it should be for low performing models. Interestingly, CoNLL seems more pessimistic than MUC for high performing models. Moreover, \newcite{barhom-etal-2019-revisiting}'s model is helped by using gold cluster for within-document entity coreference. This explains the non-zero MUC F1 performance of the baseline on the entity coreference resolution task.

Another important observation is that, when using only two embedding, the \textsc{Static/Char} model is the one experiencing the largest drop in performance (CoNLL \& Event MUC). At the same time, when using only one embedding, the \textsc{Contextual} model performs best. It even outperforms the aforementioned model with \textit{two} embeddings: \textsc{Static/Char}. These results lead us to conclude that the contextual embeddings is the most expressive for this task. This is not surprising since contextual embeddings take context into account while static and character do not. 

More interestingly, we note that removing either the static or contextual embedding  results in an average performance drop of $\sim$2.5 and $\sim$4 CoNLL points  respectively (see model \textsc{Contextual/Char} and \textsc{Static/Char}). However, when both are removed simultaneously, the performance drops by $\sim$10 CoNLL points (see model \textsc{Char}). That is, the sum of the losses incurred by removing either one of these embeddings ( $\sim$6.5) is smaller than the loss ( $\sim$10) incurred when both are simultaneously removed. Similarly, adding \textit{any one} embedding to the baseline \textsc{No word embedding} model significantly improves the latter's performance, in the range of $\sim$[+27,5 to +34,7].However, if \textit{any one} embedding is removed from the \textsc{Original} model, then the latter's performance drops by a much smaller amount,$\sim$[-1,1 to -4]. That is, removing an embedding from the \textsc{Original} model does not impact performance in a comparable way as adding an embedding to the baseline model. But performance does drop significantly when all embeddings are removed.  In other words, we face diminishing returns in terms of performance per embeddings.

\begin{figure*}[ht]
    \centering
    \includegraphics[width=\linewidth]{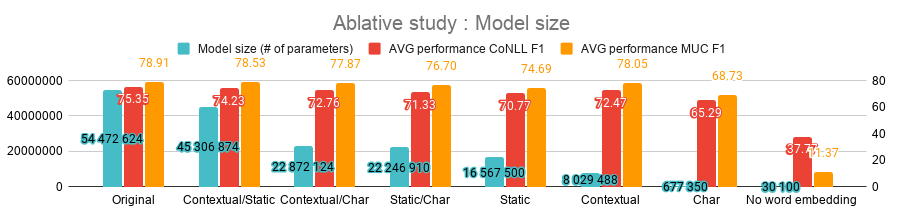}
    \caption{Comparing the size and predictive performance of the original model (using 3 embeddings) with models where we removed one, two or all three embeddings. The size of each model is the number of neural connections.}
    \label{fig:modelsize}
\end{figure*}

\subsubsection*{Impact of Dimensionality on Model Size} 

As mentioned earlier, the model size is related to the square of the input, resulting in more than 54 million parameters in the \textsc{Original} model. Thus, an important question is that of whether the gains in performance of such large models outweigh the corresponding increase in size. Our observations in  this respect are in Figure \ref{fig:modelsize}, depicting the model's respective size and predictive performance.
We observed similar diminishing returns when considering performance relative to size, i.e. increasing the model size by incorporating larger, more complex embeddings results in modest performance gains. 

The \textsc{Contextual} and \textsc{Char} models are particularly interesting. The former achieves  96\% of the performance of the \textsc{Original} model with 14.7\% of its size.
While the latter, i.e. \textsc{Char}, achieves 86\% of the performance of the \textsc{Original} model's performance, with only  1.2\% of its size. Its performance (68.43 F1) is even comparable to that of the previous event coreference resolution state-of-the-art in EvCR (69 F1) \cite{kenyon-dean-etal-2018-resolving}.

\subsubsection*{Model Size \& Run-Time}
\begin{figure}[ht]
    \centering
    \includegraphics[width=\linewidth]{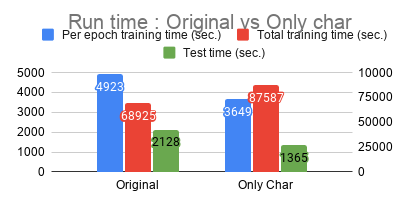}
    \caption{Run-time between the largest (54M weights) and smallest (677k weights) models. The total training time is associated with the right axis while the other measures are associated with the left axis.}
    \label{fig:runtime}
\end{figure}

Our investigations on the influence of model size on run-time and memory usage revealed paradoxical results.\footnote{Ran on a Ryzen 5 3600X CPU and a RTX 2070 Super GPU along with 32GB of RAM} They are presented in  Figure \ref{fig:runtime}. For the run-time and memory analysis, we focus only on the largest and smallest models to have a better idea of the magnitude of differences and to avoid overcrowding the Figures.

As can be seen, the huge difference in model size (54 Million vs. 0.67 Million), does not translate into equally large the differences in run-time (training \& testing) - the run-time reductions afforded by the \textsc{Char} model are relatively modest. While the actual reasons deserve further investigation, we can posit that this could be attributed to hardware and software optimization, enabling a high level of parallelization such that larger models run comparably to smaller ones.

Paradoxically, however, the larger \textsc{Original} model trains in fewer epochs than the smaller \textsc{Char} model (14 vs. 24 respectively). In consequence, it is 21\% faster to train overall (68924.8 sec. vs 87587.28 sec. or about 19h9 vs 24h19).  These results confirm the observation of \newcite{li2020train} that larger models tend to converge faster. One possible explanation could be that larger models have to optimize a error surface of higher dimensionality, leading to more possible paths for gradient descent, some of which might lead to convergence more rapidly. Thus, although adding more embedding in the model results in diminishing returns in term of predictive performance, it can lead to faster training. However, more experiments are needed to investigate this issue.

\begin{figure}[ht]
    \centering
    \includegraphics[width=\linewidth]{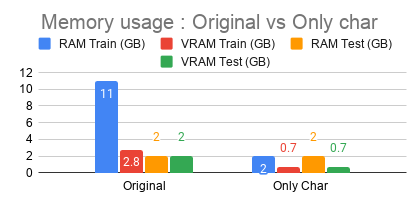}
    \caption{Memory usage between the largest (54M weights) and smallest (677k weights) models.}
    \label{fig:memusage}
\end{figure}

Concerning memory usage, we found that, as expected, the smaller \textsc{Char} model required substantially smaller amounts of memory, especially during training as evidence by Figure \ref{fig:memusage}. Note that, the RAM usage of the \textsc{Original} model is mostly due to GloVe pre-trained vectors.


\subsection{Results 2: Static Embeddings}
We now focus on  the second set of experiments, focusing our attention to static embeddings. The models concerned are from Group 2 of Table \ref{tab:models}.

\begin{figure}[th]
    \centering
    \includegraphics[width=\linewidth]{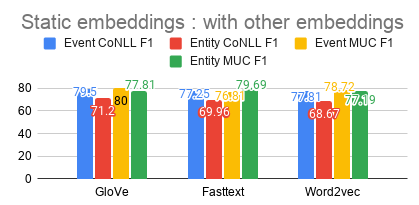}
    \caption{Comparing the predictive performance of static embeddings when used with other embeddings (ELMo and Character)}
    \label{fig:perfanal}
\end{figure}

\begin{figure}[th]
    \centering
    \includegraphics[width=\linewidth]{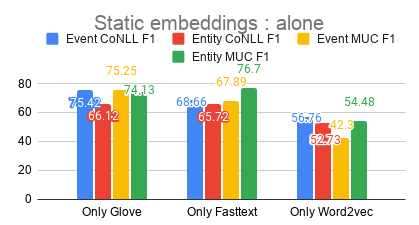}
    \caption{Comparing the predictive performance of  static embeddings when used alone}
    \label{fig:ablaperfanal}
\end{figure}

First, we varied the static embedding (GloVe, Word2Vec, FastText), while keeping the same contextual embedding and character embedding as in the \textsc{Original} model. It can be seen in Figure \ref{fig:perfanal} that, when used with other embeddings (contextual and character), all static embeddings show comparable performance. The average performance ranging from 77.12 (\textsc{GloVe}) to 75.59 (\textsc{Word2Vec}). This corroborates with our earlier findings of section \ref{ablasec} whereby the model with only contextual and character embeddings, i.e. \textsc{Contextual/char}, achieved comparable performance to the \textsc{Original} (static/contextual/char) model, indicating that the specific static embedding chosen contribute only marginally to the model's performance.

However, when used alone (see Figure \ref{fig:ablaperfanal}), we see a drastic difference in performance between them; with the average performance ranging from 72.73 (\textsc{GloVe}) to 51.56 (\textsc{Word2Vec}).

Thus, it is only when studied alone that static embeddings show their differences. Once we isolate static embeddings, we see GloVe works best for EvCR. However, for EnCR, the \textsc{Fasttext} model show significantly higher MUC. 
The better performance of GloVe and FastText with respect to word2vec can be explained by their construction. Compared to Word2Vec, GloVe takes words co-occurrence information into account. If coreferent event mentions are more likely to share co-occurring words, it would explain parts of the performance gain. FastText also outperforms Word2Vec; here the difference is that FastText takes sub-word information into account which can be advantageous for coreferent entity mentions. E.g. in Figure \ref{fig:CR_ex}, "Korea" and "Korean" have similar sub-word information.

\begin{figure}[ht]
    \centering
    \includegraphics[width=\linewidth]{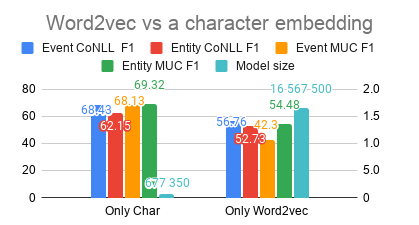}
    \caption{Comparing the predictive performance of  solely Word2Vec vs solely a character embedding}
    \label{w2vvschar}
\end{figure}

What is most surprising is that Word2Vec is significantly outperformed by a simple character embedding as we can see on Figure \ref{w2vvschar}. 
Moreover, in term of dimension Word2Vec has 300 and the character embedding has 50.
Thus, the resulting model is not only more accurate but also $\sim$24 times smaller (Figure \ref{w2vvschar}).
This could indicate that the internal structure of a word (char embedding) contains more information about possible coreferences than its usual entourage (Word2Vec). 

\subsection{Results 3: Contextual Embeddings} 
We now focus on  the third set of experiments about contextual embeddings. The models concerned are from Group 3 of Table \ref{tab:models}.

\begin{figure}[th]
    \centering
    \includegraphics[width=\linewidth]{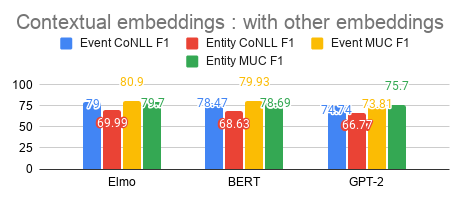}
    \caption{Comparing the predictive performance of contextual embeddings when used with other embeddings (GloVe and Character embedding)}
    \label{fig:ctxperfanal}
\end{figure}

\begin{figure}[th]
    \centering
    \includegraphics[width=\linewidth]{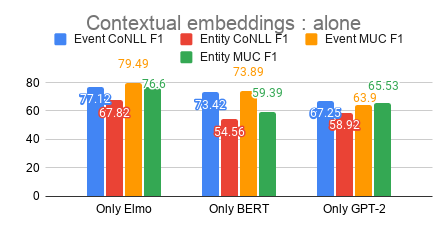}
    \caption{Comparing the predictive performance of contextual embeddings when used alone}
    \label{fig:ctxablaperfanal}
\end{figure}

Similarly to the previous section, we present the performance of different contextual embeddings when used in tandem with the static (GloVe) and character embedding of the original model (Figure \ref{fig:ctxperfanal}) or when used alone (Figure \ref{fig:ctxablaperfanal}). We see the same  as in the previous section, i.e. the difference in performance between the contextual embeddings is clearer when they are used alone versus when they are used with GloVe and a character embedding. Thus, we will only focus on the Figure \ref{fig:ctxablaperfanal} which better represent the differences between ELMo, BERT, and GPT-2.

A first observation is that BERT both outperforms and is outperformed by GPT-2 on both tasks. Specifically, BERT performs better in EvCR while GPT-2 performs better in EnCR. 

A second observation is that ELMo clearly outperforms GPT-2 and BERT on both tasks. This result contradicts \newcite{joshi-etal-2019-bert} who found that BERT  greatly outperforms ELMo on EnCR (+11.5 F1  on the GAP benchmark). Such disparity may be indicative of differences in the model and dataset. \newcite{joshi-etal-2019-bert} uses a span ranking approach which asks, for each mention, which is the most likely antecedent. This implicitly produces a tree which clusters coreferent mentions. Such method only takes local information between two mention into account while the method used in \newcite{barhom-etal-2019-revisiting} uses global information between two entity clusters and related event clusters. Moreover, ECB+ or EventCorefBank+ is an EvCR dataset first and foremost and only defines EnCR to support EvCR; you could argue that the EnCR tasks is more about argument than entities. GAP on the other hand is a corpus of ambiguous pronoun-name pairs \cite{joshi-etal-2019-bert}. 

Thus, while an EnCR task is defined by both dataset, they are significantly different. We argue that both the task definition and the use of global versus local information play a major role in the disparity between the performance reported by \newcite{joshi-etal-2019-bert} and our study.  Further confirming these findings would require evaluating \newcite{barhom-etal-2019-revisiting}'s model on GAP and \newcite{joshi-etal-2019-bert}'s on ECB+. However, these models are not interchangeable because the datasets and the task they define differs.

\section{Conclusion} 	
\label{concl}
We used the state-of-the-art in EvCR \cite{barhom-etal-2019-revisiting} as a framework to investigate the complexity-performance trade-off and compare the predictive performance of word embeddings across and within the three families.

We observed that the smallest model using solely a character embedding yielded 86\% of the performance of the original (largest) model (using Elmo, GloVe and a character embeddings) despite being only 1.2\% of its size. In fact, that smallest model achieves similar performance (68.43 F1) to the previous state-of-the-art in EvCR (69 F1) \cite{kenyon-dean-etal-2018-resolving}.

Paradoxically, we found that the largest model converged faster during training (by 21\% in overall run-time) as it took only 14 epochs vs 24 for the character model. Overall, we found size and run-time to be weakly correlated.

In addition, our experiments revealed that augmenting the model with additional embeddings does not substantially improve the performance, leading to diminishing returns in term of predictive performance per embedding.

Concerning predictive performance, one of our most interesting result is that the model using solely a character embedding significantly outperformed ($\sim$+10 F1) a larger model using solely a static embedding (Word2Vec)  while  being  radically  smaller (4\% of its size). Hence, while character embeddings have often been used as supplementary embeddings, they can actually compete with other embeddings' families in terms of predictive performance per size.

Finally, our experiments lead us to conclude that for the task of Event and Entity Coreference 
Resolution, GloVe, FastText and Elmo yielded the best predictive performance. GloVe and FastText performed best in EvCR and EnCR respectively in their family while Elmo performs best overall. 

Future directions include working on other comprehensive study of embeddings in other tasks and experimenting with CR models using different embeddings for different tasks to improve performance. E.g. GloVe and FastText  in EvCR and EnCR respectively. 

\section{Ethical considerations}
We trained 16 models over a two months period, estimated cost ranges from 350kWh to 400kWh. 
The estimated carbon impact ranges from 105Kg to 120Kg of CO2 based on local data (300g CO2/kWh).
We believe no other ethical considerations are raised by the content of this paper. 

\section*{Acknowledgments}
This research was funded by KPMG Belgium \& Luxembourg through the HEC Digital Lab/HEC-Liège/ULiège.

\bibliographystyle{META/acl_natbib}
\bibliography{META/anthology, META/bib}

\end{document}